\documentclass[sigconf,authorversion,nonacm]{acmart}

\usepackage{algorithm,algpseudocode}
\usepackage{dblfloatfix}

\AtBeginDocument{%
  \providecommand\BibTeX{{%
    \normalfont B\kern-0.5em{\scshape i\kern-0.25em b}\kern-0.8em\TeX}}}

\begin{document}

\title{An Active Learning Framework for Efficient Robust Policy Search}

\author{Sai Kiran Narayanaswami}
\authornote{Work performed when author was at the
  Robert Bosch Centre for Data Science and AI, IIT Madras.}
\affiliation{%
  \institution{The University of Texas at Austin}
    \city{}
  \country{}
}
\email{nskiran@cs.utexas.edu}

\author{Nandan Sudarsanam}
\affiliation{%
  \institution{Robert Bosch Centre for Data Science and AI\\ Indian Institute of Technology Madras}
    \city{}
  \country{}}
\email{nandan@iitm.ac.in}

\author{Balaraman Ravindran}
\affiliation{%
  \institution{Robert Bosch Centre for Data Science and AI\\ Indian Institute of Technology Madras}
    \city{}
  \country{}}
\email{ravi@cse.iitm.ac.in}

\renewcommand{\shortauthors}{Narayanaswami et. al.}

\begin{abstract}
Robust Policy Search is the problem of learning policies that do not
degrade in performance when subject to unseen environment model parameters.
It is particularly relevant for transferring policies learned in a
simulation environment to the real world. Several existing approaches
involve sampling large batches of trajectories which reflect the differences
in various possible environments, and then selecting some subset of
these to learn robust policies, such as the ones that result in the
worst performance. We propose an active learning based framework,
EffAcTS, to selectively choose model parameters for this purpose so
as to collect only as much data as necessary to select such a subset.
We apply this framework using Linear Bandits, and experimentally validate
the gains in sample efficiency and the performance of our approach
on standard continuous control tasks. We also present a Multi-Task
Learning perspective to the problem of Robust Policy Search, and draw
connections from our proposed framework to existing work on Multi-Task
Learning.
\end{abstract}





\maketitle

\footnotetext{Preprint: This is the authors' own version of their work that is published in the 5th Joint International Conference on Data Science and Management of Data (9th ACM IKDD CODS and 27th COMAD) (CODS-COMAD 2022). DOI:{10.1145/3493700.3493712} }

\section{Introduction}

Recent advances in Deep Reinforcement Learning (DRL) algorithms have
achieved remarkable performance on continuous control tasks \cite{trpo,ddpg,ppo,acer}.
Traditionally, these algorithms are used to learn policies to perform a given task in simulation. However, it has been found that policies learned in simulation often do not  perform well in, or transfer to, a real-world system that the simulation models \cite{epopt,bayessim}.
Indeed, the prospect of being able to deploy policies learned in simulation on real-world systems such as physical robots is one of the major drivers for research in Reinforcement Learning.

One class of approaches towards this goal that
has gained traction is to learn from multiple simulated domains that
approximate the real ``target'' domain\cite{epopt, dynrand, bayessim, sysid, bayessysid}. These usually correspond
to an ensemble of environment models with various parameters such as the mass
of a part of a robot or the coefficient of friction between the robot's
foot and the ground. Given such an ensemble, the problem of \emph{robust policy search}
is to learn policies that perform well across this ensemble.

One prominent group of approaches in this class involves sampling
model parameters from the ensemble and collecting batches of trajectories
simulated using these parameters \cite{ensemble-avg-wang,epopt,sysid}, which are then used
for training a policy, typically by a model-free RL algorithm.
These approaches differ mainly in the way in which they choose subsets of these trajectories
to focus on for policy learning. Although robust policy search is
inevitably a harder learning problem than standard policy search,
the amount of data collected by these methods is still quite large,
up to almost 2 orders of magnitude more than is typically required
by the usual policy optimization algorithms, regardless of the method
used to choose a subset of these trajectories for learning. Therefore, although
these approaches are shown to be effective for learning robust policies, and offer
other advantages such as reduced modeling burden, the requirement for an abundance
of data makes them computationally expensive.

In this work, we demonstrate a novel way to improve their sample complexity
while maintaining the performance and robustness of the learned policy
through the use of Active Learning for intelligently selecting model
parameters for which to sample trajectories for learning. Active Learning is used to
directly acquire some desired
subset of the trajectories (such as the subset resulting in the worst
performance), while collecting as little additional data as possible. In contrast,
existing methods sample parameters directly from the ensemble\cite{ensemble-avg-wang,epopt,sysid}, and possibly discard large portions
of the collected trajectories \cite{epopt}.
The resulting framework, EffAcTS, offers greatly improved scalability, thus
broadening its applicability to real-world problems. The structure of the framework
and the use of active learning for trajectory sampling results in some connections between 
Robust Policy Search and Multi-Task Learning. We discuss the relation between the two
problems, as well as the differences in their solution approaches.

Thus, the contributions of this paper are as follows: (1) We introduce a novel
Active Learning framework that performs more judicious collection of
trajectories for training robust policies, resulting in low sample complexity;
(2) We present an instantiation of the framework using Linear Bandits,
and perform experimental validation on environment ensembles from standard
continuous control benchmarks to empirically demonstrate significant
reductions in sample complexity while still being able to learn robust
policies; (3) We explore connections to Multi-Task Learning that are
revealed upon casting Robust Policy Search as a Multi-Task Learning
problem and discuss its relation to existing work in the area.

\section{Related Work}

\cite{ensemble-avg-wang} learn controllers with a specific functional
form using trajectories sampled for parameters drawn from an ensemble,
and optimize for the average case performance. \cite{epopt} propose
EPOpt, which learns a Neural Network (NN) policy using a model-free
DRL algorithm, but on simulated domains sampled from an ensemble of
models. An adversarial approach to training is taken that involves
selectively exposing to the model-free learner only data from those
sampled models on which the learner exhibits the least performance. EPOpt
optimizes the Conditional Value at Risk\cite{cvar-tamar}, which has also been used for
learning robust options\cite{robust-options}.
Even though this is a more sophisticated approach than the former
and is demonstrated to have greater performance and robustness, the
number of trajectories collected is still very large. A form of adversarial
training is also employed in \cite{rarl} and \cite{robustrl}, but
in these works, external (adversarial) disturbances are applied to
the agent, rather than the model itself changing.

\cite{sysid} propose an approach that optimizes the average case
performance, but additionally performs explicit system identification,
and the estimated model parameters are fed to a NN policy as additional
context information alongside the original observations. \cite{bayessim} also uses system identification on data from the real
world to decide the parameters on which to train. \cite{bayessysid}
also perform system identification, but operate in a belief space
over the model parameters. Again, the data requirements are quite
large, both for policy learning as well as system identification.

A recent work that learns from an ensemble of models is \cite{metrpo},
but the ensemble here consists of learned DNN models of the dynamics
for use in Model Based RL, rather than being induced by changing physical
properties of the environment. A similar ensemble generated by perturbing
an already learned model is used for planning through in \cite{ensemble-cio}.
This work also does not deal with model uncertainties with physical
meaning. Approaches related to learning from an ensemble of models
have also been studied under Dynamics Randomization \cite{dynrand}
and Domain Randomization \cite{domrand}.

Although \cite{epopt} uses only an appropriate subset of models to train
on, none of the above approaches consider ways to sample trajectories
only as necessary. Our proposed framework employs active learning
to decide with data from only a few model parameters the models for
which the agent requires more training. Active sampling approaches
have also been explored for task selection in Multi-Task learning
by \cite{mtrl-sharma2018}, a viewpoint we discuss in more detail
in section \ref{sec:mtrl}.

\section{Background}

\subsection{RL on an Ensemble of Models}

We work with the same setting described in \cite{epopt} where the
model ensemble is represented as a family of parametrized MDPs on
a fixed state and action space. Following the same notation, this
is the set $\mathcal{M}\left(p\right)=\left\langle \mathcal{S},\mathcal{A},\mathcal{T}_{p},\mathcal{R}_{p},\gamma,\mathcal{S}_{0,p}\right\rangle $
for each parameter $p$ in the space of parameters $\mathbb{P}$,
whose elements are respectively the state and action spaces, transition
functions, reward functions, discount factor and the initial state
distribution. Those items that are subscripted with $p$ depend on
$p$, i.e., different parameters induce different dynamics and rewards.
We note here that we say ``parameter'' even if it is a vector rather
than a real number. Further, there is a source distribution $\mathcal{P}$
that indicates the likelihood of any particular $p\in\mathbb{P}$
in the model ensemble.

We denote the typical trajectory from any of these MDPs by $\tau=\left\{ s_{t},a_{t},r_{t}\right\} _{t=0}^{T}$,
where $T$ is the time horizon, and the discounted return from the
start state $R\left(\tau\right)=\sum_{t=0}^{T}\gamma^{t}r_{t}$. These
trajectories are generated by following a policy which are parameterized
by a vector $\theta$, which we denote by $\pi_{\theta}$. We define
the performance at parameter $p$ as the expected discounted return
from the start state in $\mathcal{M}\left(p\right)$

\begin{equation}
\eta\left(\theta,p\right)=\mathbb{E}_{\tau}\left[\sum_{t=0}^{T}\gamma^{t}r_{t}\,\Bigg|\,p\right]\label{eq:perf_definition}
\end{equation}

% \subsection{Robust Policy Search and EPOpt}
\subsection{Robust policy learning via CVaR optimization}

Robust Policy Search seeks policies that perform well across all parameters
in $\mathbb{P}$, and do so without knowing the parameter for the
MDP on which they are being tested. This translates to being able
to perform well on some unknown target domain, and also potentially
handle variations not accounted for in $\mathbb{P}$. The intuitive
objective for this is to consider the average performance of the policy
over the source distribution $\eta_{D}\left(\theta\right)=\mathbb{E}_{p\sim\mathcal{P}}\left[\eta\left(\theta,p\right)\right]$.
However, this objective could be close to the maximum even if there
are sharp drops in performance in some regions of $\mathbb{P}$. A different
objective used by approaches such as \cite{epopt} is the Conditional Value at Risk (CVaR) formulation from
\cite{cvar-tamar}, which considers the performance across only the
subset of $\mathbb{P}$ that corresponds to the bottom $\epsilon$
percentile of returns from $\mathcal{P}$, for a given $\epsilon \in (0,1]$. This has the effect that
policies which have such sharp drops in performance (i.e bad worst
case performance) are no longer considered good solutions.

\subsection{Linear Stochastic Bandits}

Here, we provide a quick overview of Linear Stochastic Bandits (LSB)
since they play an important role as solutions to the Active Learning
problem in section \ref{sec:eats}.

The LSB problem is one of finding the optimal arm from a given set
of arms $\mathcal{X}$ similar to the standard Multi-Armed Bandit
(MAB) problem, but with the average reward from each arm being an
unknown linear function of the features associated with that arm.
That is, if $x\in\mathcal{X}$ is an arm, and we also denote its features
by $x$, the reward is given by $r\left(x\right)=x^{T}\theta^{*}+\xi$,
where $\xi$ is some zero-mean noise, and $\theta^{*}$ gives the
parameters for said linear function. Thus, finding the optimal arm
amounts to estimating $\theta^{*}$. Although it may seem restrictive
to assume a linear dependence on the arms, more expressiveness
can be achieved by using a feature transformer, in a manner similar
to the practice for Linear Regression. The feature transformer is a
function $f:\mathcal{X}\rightarrow \mathbb{R}^{d}$ that maps each arm
$x\in\mathcal{X}$ to a feature vector $f(x)$, and the LSB learner
would be estimating $\theta^{*}$ so that $r\left(x\right)=f(x)^{T}\theta^{*}+\xi$.

There have been several approaches to solving the LSB problem under
various objectives. One group of works \cite{oful,linucb} are based
on the principles of the Upper Confidence Bound Algorithm for MAB
problems. The other popular class of approaches \cite{lsb-thompson-revisited,lsb-thompson-ctx}
is based on Thompson Sampling.

\section{Active Learning for Efficient Trajectory Sampling\label{sec:eats}}

Existing approaches for Robust Policy Search are based on collecting
trajectories at various parameters from $\mathbb{P}$, and using
some subset of these trajectories to optimize the policy being learned.
As mentioned before, the number of trajectories required can be very
large, up to 2 orders of magnitude more than required for a standard
RL problem. Although Robust Policy Search is expected to require more
data than standard RL, improvements to its sample efficiency are still
necessary in order for it to be viable in complex real-world systems.

To motivate our developments to improve on the sample efficiency,
we start with the observation that there is some functional dependence
of the performance of a given policy on the model parameter corresponding
to a task under consideration. Existing approaches 
all disregard this dependence when evaluating their objective,
leading to increased sample complexity. The increase in sample complexity is
more severe when using the CVaR objective such as in \cite{epopt}, due to
having to discard most of the trajectories collected in order to estimate the CVaR.
In fact, for a standard value of $\epsilon=0.1$, 90\% of the collected
trajectories need to be discarded (more generally, a $1-\epsilon$ fraction). Here, we wish to devise a strategy to
utilize the information from the aforementioned functional dependence
effectively so as to minimize such wastage.

% The CVaR objective estimation process
% of EPOpt disregards this functional dependence as it does not assume
% such a dependence exists (in the sense that even if the performance
% was completely independent of the model parameter, it would still
% be able to come up with a sample from the bottom $\epsilon$ percentile
% of trajectories). Thus, an unavoidable side effect of this process
% is that EPOpt discards many of the trajectories it collects (for $\epsilon=0.1$,
% 90\% of the collected trajectories are discarded). 

\subsection{Active Learning and the EffAcTS framework}

\begin{figure}[tb]
\begin{centering}
\includegraphics[width=\columnwidth]{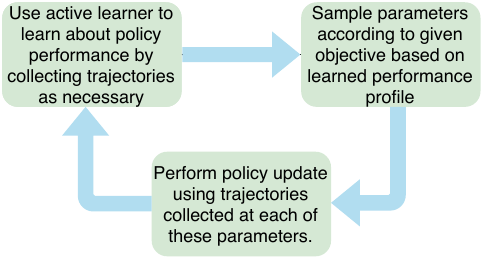}
\par\end{centering}
\caption{Main loop of the EffAcTS framework.}

\label{EffAcTS-diagram}\end{figure}%

Active Learning is a paradigm where the agent chooses data to learn
from based on its previous experience (see \cite{al-survey} for a
comprehensive survey). It has been used to speedup learning tasks,
especially in situations with limited data. An active learner not
only needs to work with as few samples as possible, it also needs
to account for the uncertainty in whatever data it has collected.
These are exactly the desiderata of the required strategy, as it must
be able to fit the performance function across $\mathbb{P}$ by collecting
as few trajectories as possible, which come with noisy evaluations
of the performance. Thus, quite clearly, the problem of efficiently
performing such sampling is connected to active learning. The case
of active learning that is of interest to us is when the agent is
allowed to sample output for arbitrary points in the input space (instead
of having to choose points from a finite dataset). The input here
is some parameter $p$ from $\mathbb{P}$, and the output is the return
from one trajectory collected at $p$.

We now outline our active learning framework for learning robust policies.
Each policy update involves selectively generating trajectories to
be sent to the batch policy optimization algorithm. This is done in
two phases as follows:

\paragraph*{Performance Assessment:}

In the first phase, active learning is used to assess the performance
of the current policy. To do this, an active learner sequentially
picks some parameters and trajectories are sampled for each of them
by setting the environment to these parameters and running the current
policy. After a particular number of trials, it is in theory expected
to have a reasonably good approximation of the performance as a function
of the parameters, say $\hat{\eta}_{\theta_{i}}(p)$. Note that we
have used $\theta_{i}$ as a subscript since this function is specific
to the current policy with parameters $\theta_{i}$.

The active learner and this process are encapsulated in a subroutine
\textsc{LearnPerf} that takes as input some policy parameters, $\theta$
and returns the function $\hat{\eta}_{\theta}$ as described above.
\textsc{LearnPerf} could be equipped with persistent memory (e.g.
to store data from previous iterations), and no assumptions are made
about the form of the output $\hat{\eta}_{\theta}$ other than that
it can be evaluated at any given $p\in\mathbb{P}$.

\paragraph*{Parameter Selection:}

The second phase uses this assessment to decide which parameters from
$\mathbb{P}$ trajectories need to be collected for. This selection
is guided by some given objective that enforces robustness. Another
subroutine, \textsc{SelectParams} carries out this selection, taking
as input a performance profile $\hat{\eta}_{\theta}$. Based on this
performance profile, it returns a set $\mathbf{P}$ of parameters
at which the policy needs to be trained.

The policy update is then performed using trajectories collected for
each parameter in $\mathbf{P}$, and this is done for a given number
($N_{iters}$) of iterations. We call the resulting framework EffAcTS
(Efficient Active Trajectory Sampling), and summarize it in Algorithm
\ref{effacts-algo} and Figure \ref{EffAcTS-diagram}. Any particular
instantiation of EffAcTS is defined by the choice of active learning
algorithm and also the scheme used to select which parts of $\mathbb{P}$
to sample from, i.e by specifying the \textsc{LearnPerf} and \textsc{SelectParams}
subroutines.

\begin{algorithm}[tb]
\begin{centering}
\begin{algorithmic}[1]
\For{$i=0\ldots N_{iters}-1$}
 \State$\hat{\eta}_{\theta_{i}}\leftarrow$ \textsc{LearnPerf}($\theta_{i}$)
\State $\mathbf{P}\leftarrow$ \textsc{SelectParams}($\hat{\eta}_{\theta_{i}}$)
\State $\boldsymbol{\tau}\leftarrow\left\{ \right\} $
\For{each $p\in\mathbf{P}$}
\State $\tau_{p}\leftarrow$ Trajectory collected at $p$ using $\pi_{\theta_{i}}$
\State $\boldsymbol{\tau}\leftarrow\boldsymbol{\tau}\cup\tau_{p}$
\EndFor

\State$\theta_{i+1}\leftarrow$\textsc{BatchPolOpt}($\boldsymbol{\tau}$, $\theta_{i}$)
\EndFor

\State \Return $\theta_{N_{iters}}$
\end{algorithmic}
\end{centering}
\caption{The EffAcTS framework}
\label{effacts-algo}
\end{algorithm}%

\subsection{Applying EffAcTS}

We analyze one such instantiation based on the CVaR objective for
parameter selection. To generate samples of the bottom $\epsilon$
percentile of trajectories, a batch of parameters is sampled from
$\mathcal{P}$, and trajectories are collected only for those that
are in the worst $\epsilon$ percentile of performance according to
$\hat{\eta}_{\theta}$. The use of Bandit algorithms for active learning
is well studied in both Multi-Armed Bandit \cite{al-mab,al-ucb-mab}
as well as Linear Stochastic Bandit \cite{al-lsb} settings. The
parameter spaces involved in Robust RL are invariably continuous.
LSBs and Gaussian Process Regression are two well-known approaches that can perform
active regression on continuous spaces. LSBs, however, are specialized
in the sense that they quickly seek out areas that lead to high reward,
a property that we make use of as described shortly.
Considering that Gaussian
Processes are more computationally expensive, and also that LSBs are
simpler to implement and can be
very efficient, especially when data is scarce, we turn to LSBs as
the active learner in our experiments.

% Since
% the parameter space is invariably continuous, we turn to LSBs as the
% active learner in our experiments.

Its arms are simply a large enough collection of parameters spread
across $\mathbb{P}$. Pulling an arm $p\in\mathbb{P}$ results in
a trajectory being sampled at $p$. Feedback is given to the bandit
in an adversarial manner, being proportional to the negative of the
return obtained on that sampled trajectory. This causes it to seek
out regions with low performance, and in the process learn about the
performance across $\mathcal{P}$. This behaviour is especially useful
for identifying the worst-case parameters that are used in optimizing
the CVaR objective.

We note that although the bandit's
learning phase is inherently serial, it is still possible to collect
the trajectories for the estimated worst $\epsilon$ percentile of
parameters in parallel.

We call this algorithm EffAcTS-C-B and its \textsc{LearnPerf} and
\textsc{SelectParams} subroutines described in Algorithm \ref{e2po_trj}.
The following hyperparameters are introduced: $N_{B}$, the number of trajectories sampled by the bandit
in the course of its learning, $N_{C}$, the total number of parameters
chosen (this means that $\left\lceil \frac{N_{C}}{\epsilon}\right\rceil $
parameter values are drawn from the source distribution and their
performance is estimated using the bandit, but only the bottom $N_{C}$
of those are used to collect trajectories). The most critical component
is the LSB learner $B$ which incorporates internally a feature transformer
$f$ that takes in a parameter from the source distribution's support
and applies some transformation on it (including possibly standardization),
and also scales the negative returns given to it appropriately.

\begin{algorithm*}[tb]
\begin{minipage}[t]{0.5\textwidth}%
\begin{algorithmic}[1]

\Function{\textsc{LearnPerf}}{$\theta$}
\For{$i=1\ldots N_{B}$}
\State{$p_{i}\leftarrow$ Optimal arm estimate by $B$.}
\State{$\tau_{i}\leftarrow$ %
\begin{minipage}[t]{0.5\textwidth}%
Trajectory collected at $p$ using $\pi_{\theta}$%
\end{minipage}}
\State{$R_{i}\leftarrow$ Return from $\tau_{i}$}
\State{Update $B$ with $\left(p_{i},-R_{i}\right)$}
\EndFor
\State{$\hat{\eta}_{\theta}\leftarrow$ %
\begin{minipage}[t]{0.65\textwidth}%
Function parameterized by $B$'s current weights%
\end{minipage}}
\State \Return $\hat{\eta}_{\theta}$
\EndFunction

\end{algorithmic}%
\end{minipage}%
\begin{minipage}[t]{0.5\textwidth}%
\begin{algorithmic}[1]

\Function{\textsc{SelectParams}}{$\hat{\eta}_{\theta}$}
\State{$\mathbf{P}\leftarrow\left\{ \right\} $, $\hat{\mathbf{R}}\leftarrow\left\{ \right\} $}
\For{ $j=1\ldots\left\lceil \frac{N_{C}}{\epsilon}\right\rceil $}
\State{$p_{j}\leftarrow$ Parameter sampled from $\mathcal{P}$}
\State{$\hat{R}_{j}\leftarrow$ $\hat{\eta}_{\theta}(p_{j})$}
\State{Add $\hat{R}_{j}$ to $\hat{\mathbf{R}}$ and $p_{j}$ to
$\mathbf{P}$}
\EndFor
\State{$\mathbf{P}_{C}\leftarrow$%
\begin{minipage}[t]{0.65\textwidth}%
Subset of $\mathbf{P}$ giving rise to the bottom $\epsilon$ percentile
of returns in $\hat{\mathbf{R}}$%
\end{minipage}}
\State \Return $\mathbf{P}_{C}$
\EndFunction

\end{algorithmic}%
\end{minipage}

\caption{Definitions of \textsc{LearnPerf} and \textsc{SelectParams} subroutines
for EffAcTS-C-B}

\label{e2po_trj}

\end{algorithm*}

\subsection{Sample Efficiency}

Due to the fact that EffAcTS-C-B strategically chooses parameters at which
to collect trajectories, we expect it to be able to maintain robustness and
performance while collecting fewer samples than existing approaches. Consider
for instance EPOpt\cite{epopt}, which discards a $1-\epsilon$ fraction of the trajectories
it collects. If EPOpt collects $N$ trajectories, and EffAcTS-C-B's
bandit learner is allowed $N_{B}$ ``arm pulls'', with the same
number of trajectories being used for learning as EPOpt (i.e $N_{C}=\epsilon N$),
the ratio of the total amount of data collected by the two algorithms is $\left(\frac{N_{B}+\epsilon N}{N}\right)$.
For a nominal setting of $N=240$, $\epsilon=0.1$ and $N_{B}=24$,
this results in a dramatic 80\% reduction in the amount of data collected.
We later show that efficiency gains of this order are indeed attainable
in practice.

\begin{table*}[!tb]
\begin{centering}
\begin{minipage}[t]{0.5\textwidth}%
\begin{center}
\begin{tabular}{ccccc}
\toprule 
Parameter & $\mu$ & $\sigma$ & Low & High\tabularnewline
\midrule
\midrule 
Mass & 6.0 & 1.5 & 3.0 & 9.0\tabularnewline
\midrule 
Friction & 2.0 & 0.25 & 1.5 & 2.5\tabularnewline
\midrule 
Damping & 2.5 & 2.0 & 1.0 & 4.0\tabularnewline
\midrule 
Inertias & 1.0 & 0.25 & 0.5 & 1.5\tabularnewline
\bottomrule
\end{tabular}
\par\end{center}%
\end{minipage}%
\begin{minipage}[t]{0.5\textwidth}%
\begin{center}
\begin{tabular}{ccccc}
\toprule 
Parameter & $\mu$ & $\sigma$ & Low & High\tabularnewline
\midrule
\midrule 
Mass & 6.0 & 1.5 & 3.0 & 9.0\tabularnewline
\midrule 
Friction & 0.5 & 0.1 & 0.3 & 0.7\tabularnewline
\midrule 
Damping & 1.5 & 0.5 & 0.5 & 2.5\tabularnewline
\midrule 
Inertias & 0.125 & 0.04 & 0.05 & 0.2\tabularnewline
\bottomrule
\end{tabular}
\par\end{center}%
\end{minipage} 
\par\end{centering}
\caption{Description of the source domain distribution for the Hopper (Left)
and Half-Cheetah (Right) tasks. The values given here specify the
means ($\mu$) and standard deviations ($\sigma$) for normal distributions
truncated at the low and high points mentioned here. This is equivalent
to the probability density of a normal distribution with parameters
($\mu,\sigma$) being zeroed outside the interval {[}Low, High{]},
and being normalized so that it integrates to 1.}

\label{param_dist}
\end{table*}

\section{Connections to Multi-Task Learning\label{sec:mtrl}}

The problem of robust policy search on an ensemble of models can also
be viewed as a form of transfer learning from simulated domains to
an unseen real domain (possibly without any training on the real domain,
which is referred to as direct-transfer or jumpstart \cite{jumpstart}).
Further, the process of learning from an ensemble of models can be
viewed as a Multi-Task Learning (MTL) problem with the set of tasks
corresponding to the set of parameters that constitute the source
domain distribution. Learning a robust policy corresponds to maintaining
performance across this entire set of tasks, as is usually the goal
in MTL settings. MTL, which is closely related to transfer learning,
has been studied in the DRL context in a number of recent works \cite{poldist,prognets,actormimic,mtrl-sharma2018}.
However, these works consider only discrete and finite task sets,
whereas model parameters form a (usually multi-dimensional) continuum.
More generally, we can think of MTL with the task set being generated
by a set of parameters, and refer to such problems as \emph{parameterized
MTL} problems, with Robust Policy Search being an instance of this
setting.

\cite{mtrl-sharma2018} has employed a bandit based active sampling
approach similar to what we have described here to intelligently sample
tasks (from a discrete and finite task set) to train on for each iteration.
The feedback to the bandit is also given in an adversarial manner.
However, we note several differences when it comes to a parameterized
MTL setting. The first is the functional dependence of the task performance
to an underlying parameter as discussed earlier. In the discrete MTL
settings usually studied (such as Atari Game playing tasks), there
is no such visible dependency that can be modeled. This means that
any algorithm that is adapted to the parameterized setting need
to be reworked to utilize such dependencies as EffAcTS does. The task
selection procedure in EffAcTS differs from the one in \cite{mtrl-sharma2018}
in that the performance is sampled for several tasks in between iterations
of policy training. Additionally, one single task is not chosen in
the end, rather the active learner is used to inform the selection
of a group of tasks as necessary for the given objective.

\section{Experiments}

\begin{figure*}[tb]
\begin{centering}
\includegraphics[width=0.5\textwidth]{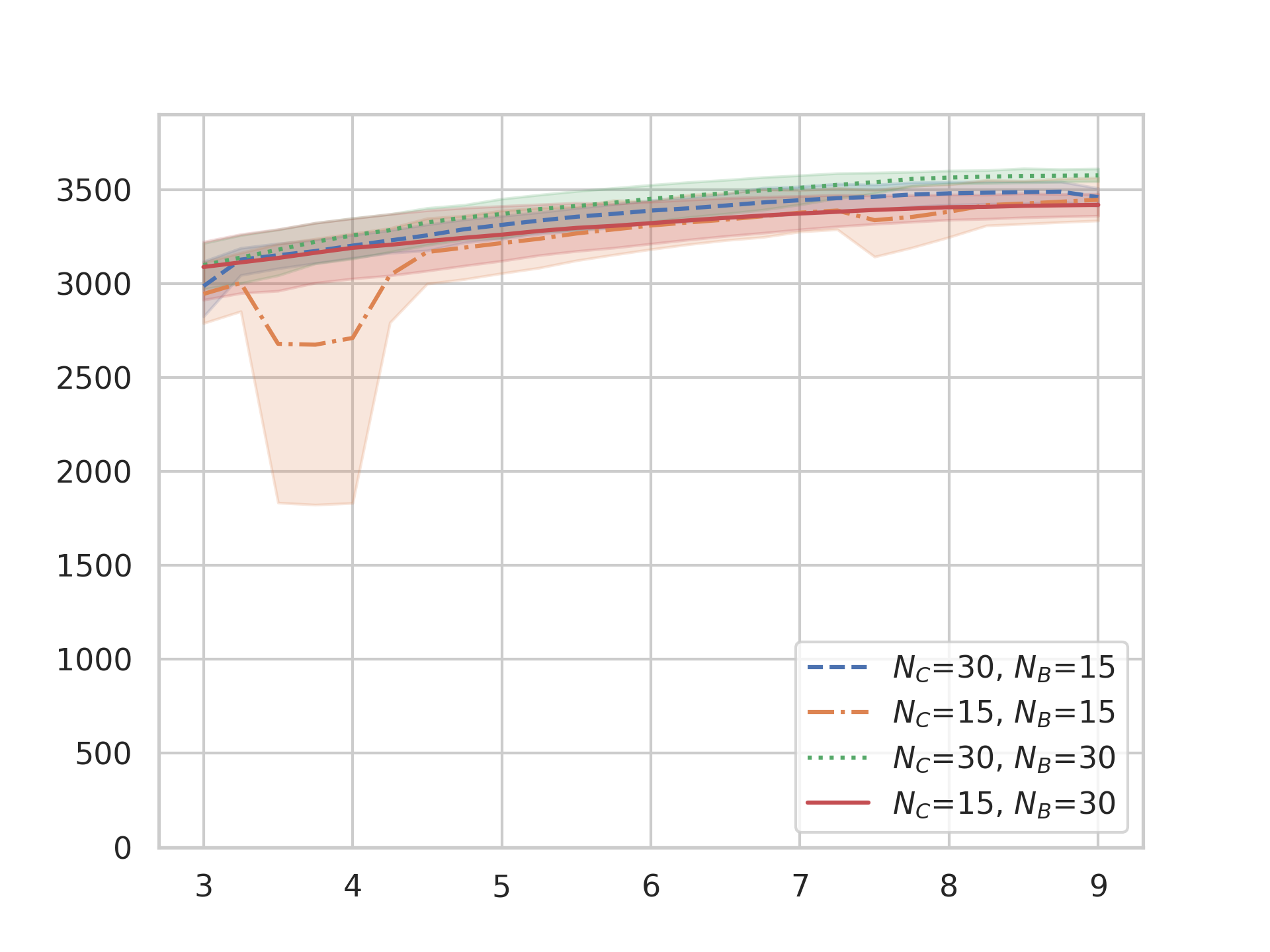}\includegraphics[width=0.5\textwidth]{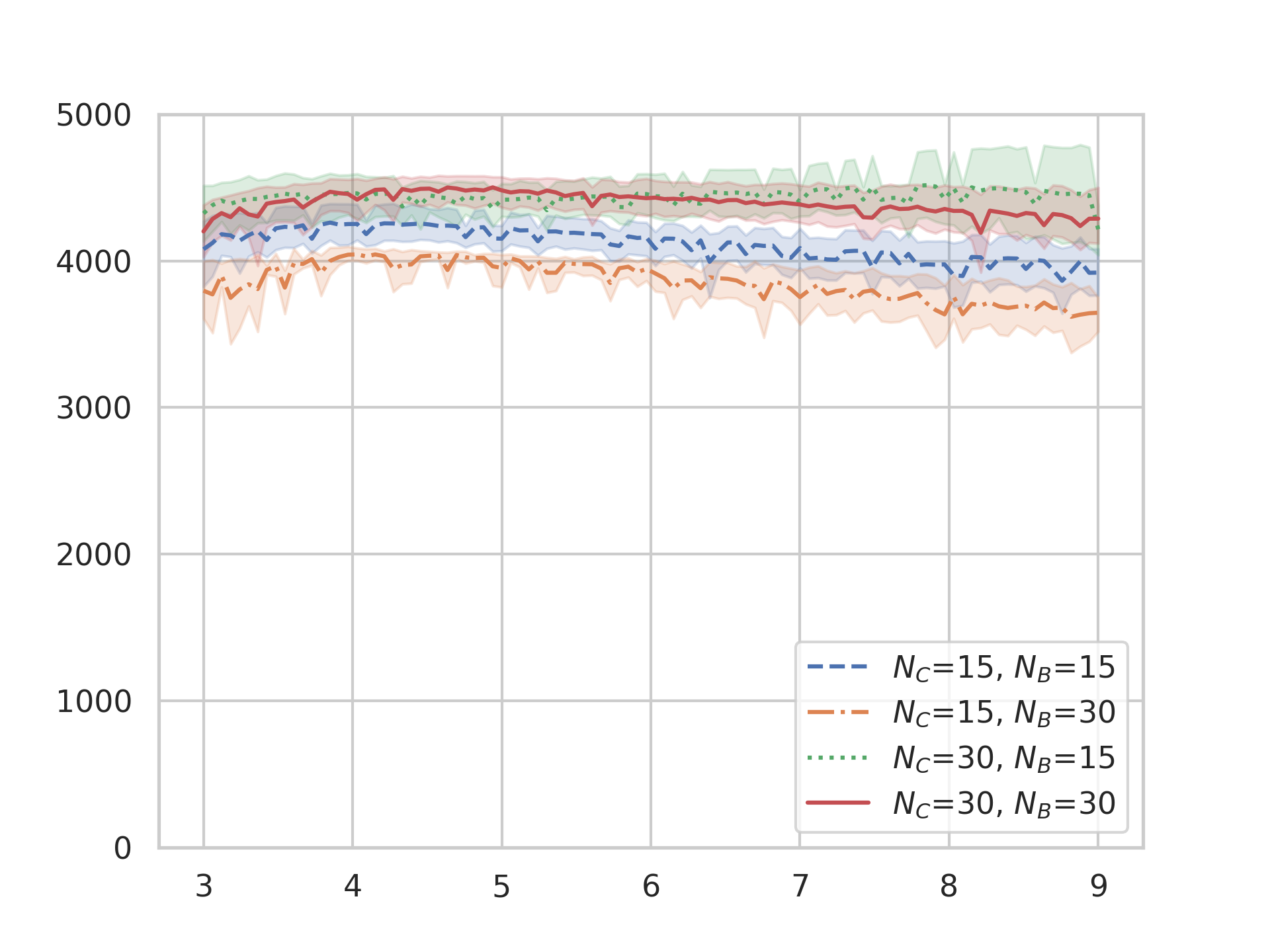}
\par\end{centering}
\caption{Performance (y-axis) as a function of torso mass (x-axis) for $\epsilon=0.1$ for the
Hopper (Left) and Half-Cheetah (Right) tasks. The bands indicate the
confidence intervals of the performance measured across 5 runs of
the entire training procedure. All of the configurations tested use at least 75\%
fewer trajectories than EPOpt, and with the exception of $N_{C}=15$ and $N_{B}=15$
in the Hopper task alone,
perform to the same extent and are as robust as EPOpt (acccording to results
reported in \cite{epopt}).}

\label{plot-1d-perf}
\end{figure*}

As EPOpt\cite{epopt} uses the CVaR objective, it is a very suitable baseline against
which to compare EffAcTS-C-B for demonstrating the benefits of introducing
the active learner. We conduct experiments to answer the following
questions, which we will reference as RQ1, RQ2 etc. in later discussions:
\begin{enumerate}
\item Do the policies learned using EffAcTS-C-B suffer any degradation in
performance from that of EPOpt? Is robustness preserved across the
same range of model parameters as in EPOpt?
\item Does the bandit active sampler identify with reasonable accuracy the
region corresponding to the worst $\epsilon$ percentile of performance,
and does it achieve a reasonable fit to the performance across the
range of parameters (i.e has it explored enough to avoid errors due
to the noisy evaluations it receives)?
\item How much can the sample efficiency be improved upon? This mainly boils
down to asking how few trajectories are sufficient for the bandit
to learn well enough.
\end{enumerate}

\subsection{Implementation Details and Hyperparameters}

The experiments are performed on the standard Hopper and Half-Cheetah
continuous control tasks available in OpenAI Gym \cite{gym}, simulated
with the MuJoCo Physics simulator \cite{mujoco}. As in \cite{epopt},
some subset of the following parameters of the robot can be varied:
Torso Mass, Friction with the ground, Foot Joint Damping and Joint
Inertias. We also use the same statistics for the source distributions
of the parameters which are described in Table \ref{param_dist}.

We emphasize that because we are using the same environments in \cite{epopt}
and also the same policy parameterization, results reported there
can be used directly to compare against EffAcTS-C-B.

\begin{table*}[!t]
\begin{centering}
\begin{tabular}{ccccc}
\toprule 
Hyperparameters & Median \%tile & Avg \%tile & Std. Dev. & Max \%tile\tabularnewline
\midrule
\midrule 
$N_{C}$=15, $N_{B}$=15 & 14.6 & 15.9 & 7.7 & 32.2\tabularnewline
\midrule 
$N_{C}$=15, $N_{B}$=30 & 14.9 & 23.4 & 24.0 & 98.0\tabularnewline
\midrule 
$N_{C}$=30, $N_{B}$=15 & 13.7 & 23.0 & 19.4 & 74.7\tabularnewline
\midrule 
$N_{C}$=30, $N_{B}$=30 & 11.6 & 14.9 & 10.6 & 47.8\tabularnewline
\bottomrule
\end{tabular}
\par\end{centering}
\caption{Statistics for the percentiles described in section \ref{subsec:Analysis-of-Bandit}
for the Hopper task's 1-D model ensemble with $\epsilon=0.1$, which
are measured every fifth iteration from the 100th to the 150th iterations. }

\label{pctl-table}
\end{table*}

We use the hyperparameter settings for TRPO suggested in
OpenAI Baselines \cite{baselines} on which our implementation is
also based. These are shown in Table \ref{trpo-hyp}.

\begin{table}[h]
\begin{centering}
\begin{tabular}{|c|c|}
\hline 
Hyperparameter & Value\tabularnewline
\hline 
\hline 
Timesteps per batch & 1024\tabularnewline
\hline 
Max KL & 0.01\tabularnewline
\hline 
CG Iters & 10\tabularnewline
\hline 
CG Damping & 0.1\tabularnewline
\hline 
$\gamma$ & 0.99\tabularnewline
\hline 
$\lambda$ & 0.98\tabularnewline
\hline 
VF Iterations & 5\tabularnewline
\hline 
VF Stepsize & 1e-3\tabularnewline
\hline 
\end{tabular}
\par\end{centering}
\caption{Hyperparameters for TRPO}
\label{trpo-hyp}

\end{table}

One difference from EPOpt's implementation is that we use
Generalized Advantage Estimation \cite{gae} instead of subtracting
a baseline from the value function. For value function estimation
using a critic, we use the same NN architecture as the policy, 2 hidden
layers of 64 units each. Policies are parameterized
with NNs and have two hidden layers with 64 units each, and use $\tanh$
as the activation function.

\subsection{Bandit Algorithm}

For all our experiments, we implement the bandit learner using Thompson
Sampling due to its simplicity, following the version described in
\cite{lsb-thompson-revisited}. The hyperparameters
introduced by this are as in Table \ref{bandit-hyp} (they have the same names as in
the paper).

\begin{table}[tb]
\begin{centering}
\begin{tabular}{|c|c|}
\hline 
Hyperparameter & Value\tabularnewline
\hline 
\hline 
$R$ & 5.0\tabularnewline
\hline 
$\delta$ & 0.1\tabularnewline
\hline 
$\lambda$ & 0.5\tabularnewline
\hline 
\end{tabular}
\par\end{centering}
\caption{Thompson Sampling Hyperparameters}
\label{bandit-hyp}

\end{table}

The first two parameters control the amount of exploration performed
during Thompson Sampling, while $\lambda$ is a regularization parameter
for the parameter estimates.

The arms of the bandit are model parameters taken uniformly across
the domain and converted to feature values. In order to allow for some degree
of expressiveness for the fit, we apply polynomial transformations
of some particular degree to the model parameters. The features input to
the bandit are 4th degree polynomial terms generated from the model
parameters. This amounts to 5 terms in the 1-D case and 15 in the
2-D case. These arm ``representations'' are then standardized before
being used by the bandit. The negative returns given as feedback to
the bandit are scaled by a factor of $10^{-3}$.

\subsection{(RQ1) Performance and Robustness}

In this section, we perform the following experiment to evaluate
EffAcTS-C-B for the objectives of RQ1. In this experiment, only one
environment parameter (torso mass) is varied, creating a 1-D model ensemble
on which to evaluate the algorithm.

The torso mass is varied in both the Hopper and the Half-Cheetah
domains keeping the rest of the parameters fixed at their mean values.
The performance of the EffAcTS-C-B learned policy is then tested across
this range, and the results are shown in Figure \ref{plot-1d-perf}.
We use Trust Region Policy Optimization (TRPO) \cite{trpo} for batch
policy optimization and run it for 150 iterations. For this part,
we use 4th degree polynomial transformations.

We see that the policy is indeed robust as it maintains its performance
across the range of values of the torso mass, and it achieves near
or better than the best performance for both tasks as reported in
\cite{epopt} in all but one case, with $N_{C}=15$ and $N_{B}=15$
for Hopper. This setting samples the least number of trajectories
per iteration, 30, which is just one eighth of the 240 drawn in EPOpt.
Although there is one region where it is unstable, it is still able
to maintain its performance everywhere else, thus attaining
the same level of robustnes as EPOpt. Further, the performance achieved
is almost as good as, and possibly better than EPOpt\cite{epopt}.

For the other settings which use more trajectories, this does not
happen, and even at $N_{C}=30$ and $N_{B}=30$ which samples the
most trajectories, a 75\% reduction in samples collected is achieved
over EPOpt (the total number of iterations is the same). The other
two settings which perform almost as well in both tasks collect 45
each, which amounts to an even larger reduction of 81.2\%. In the
case of $N_{C}=15$ and $N_{B}=15$ with the Half-Cheetah Task, this
number is pushed even further to 87.5\% while still retaining performance
and robustness.

% In Appendix \ref{sec:results-2d}, we also present results when vary
% the friction with the ground in addition to the torso mass, thus creating
% a two dimensional ensemble of parameters. Again, full performance
% is maintained over almost all of the parameter space in both domains,
% being comparable to or better than in \cite{epopt}.

\subsection{(RQ1) Performance on a 2-D Model Ensemble\label{sec:results-2d}}

\begin{figure}[tb]
\begin{centering}
\includegraphics[width=0.5\columnwidth]{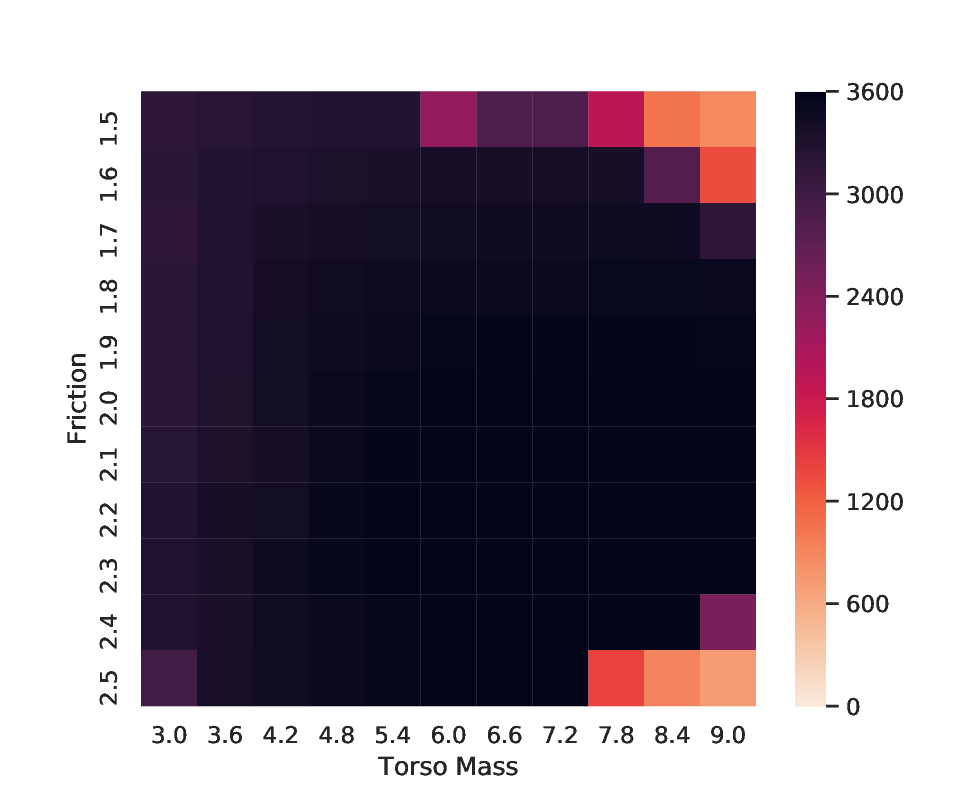}\includegraphics[width=0.5\columnwidth]{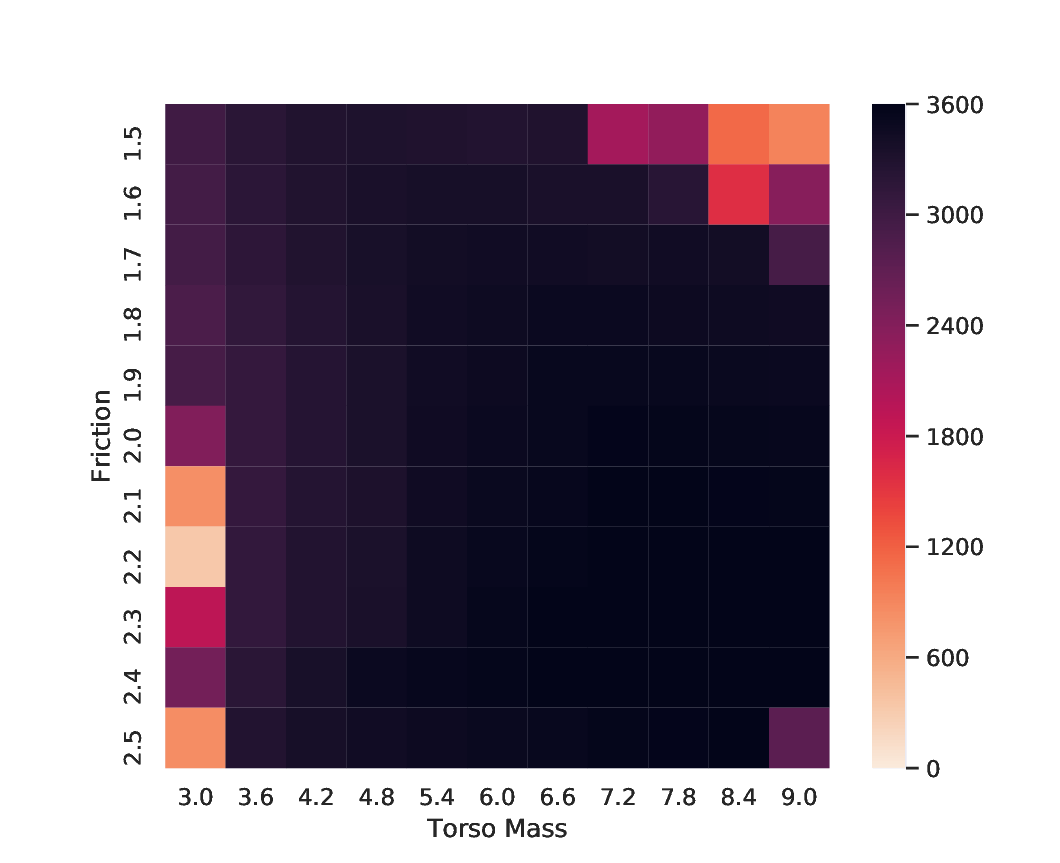}
\par\end{centering}
\begin{centering}
\includegraphics[width=0.5\columnwidth]{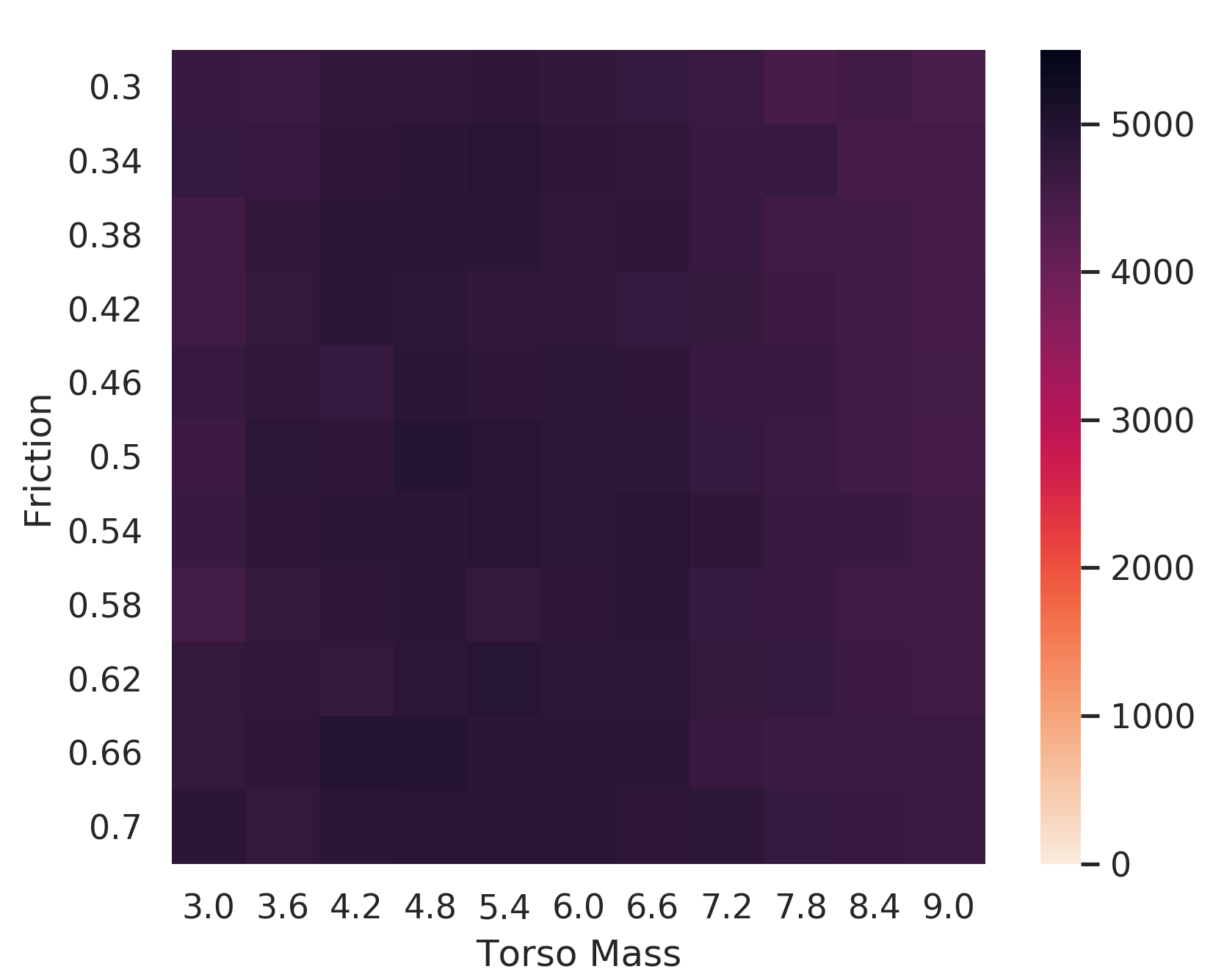}\includegraphics[width=0.5\columnwidth]{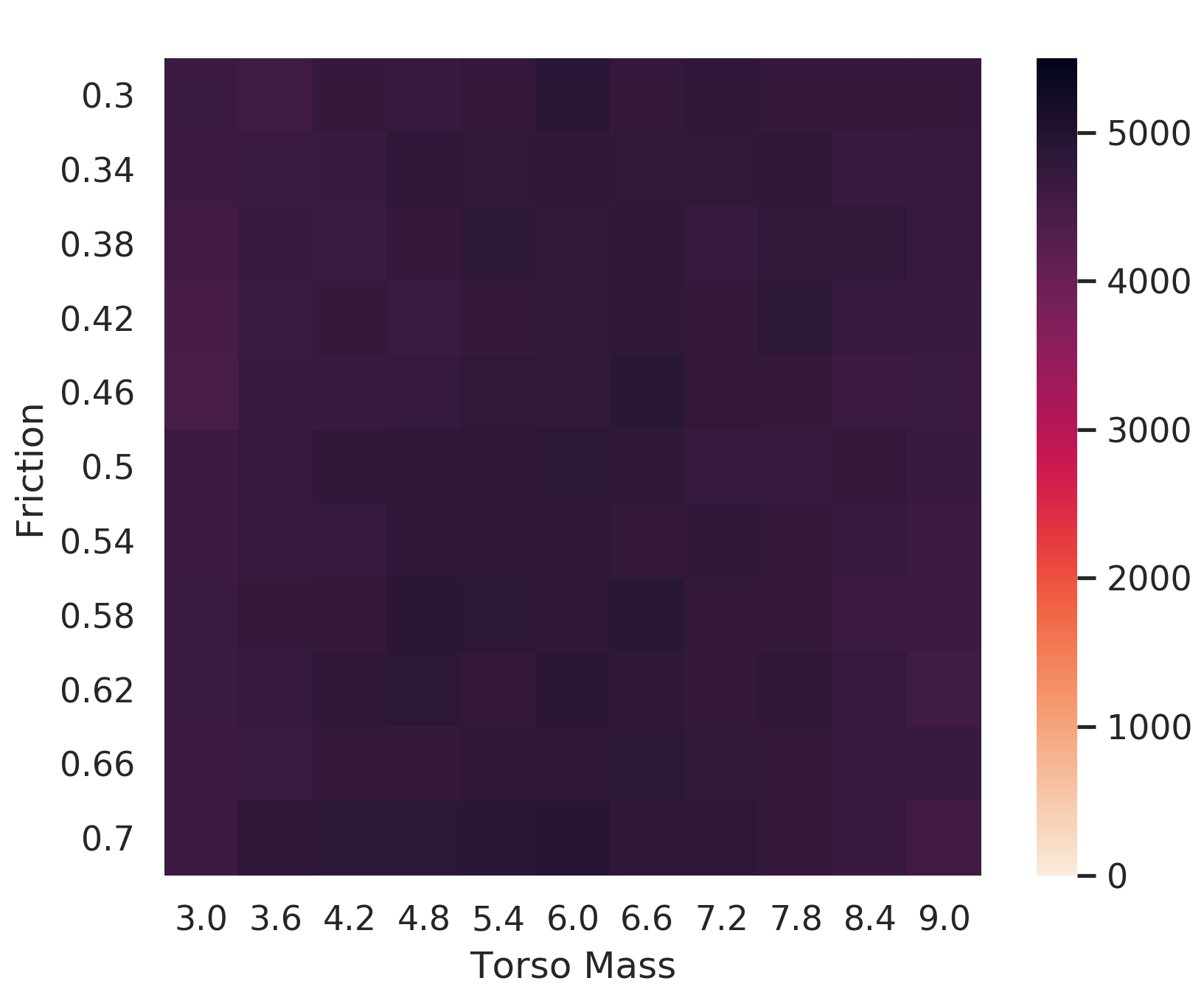}
\par\end{centering}
\caption{Performance as a function of torso mass and ground friction on the
Hopper (left half) and Half-Cheetah (right half) tasks for $\epsilon=0.1$
for one run or EffAcTS-C-B. In each row, Left: $N_{C}$=30, $N_{B}$=30.
Right: $N_{C}$=30, $N_{B}$=50.}

\label{hopper-2d-perf}
\end{figure}

In this question, we further investigate the scalability of the algorithm
to larger model ensembles, again in the context of RQ1. In this experiment,
we vary the friction with the ground in addition to the torso mass,
thus creating a two dimensional ensemble of parameters. Here, we run
TRPO for 200 iterations, and again use 4th degree polynomial transformations.
Figure \ref{hopper-2d-perf} shows the results obtained.

Full performance is maintained over almost all of the parameter space
in both domains, again being comparable to or better than in \cite{epopt}.
Notably, with $N_{C}$=30, $N_{B}$=30, the same 75\% reduction in
collected trajectories is obtained even in a higher dimensional model
ensemble. This is also despite the added challenge of an increased
number of parameters for the bandit to fit (15 as opposed to 5 in
the previous experiment).

\subsection{Visualizing the Bandit Active Learner}

\begin{figure}[tb]
\begin{centering}
\includegraphics[width=0.9\columnwidth]{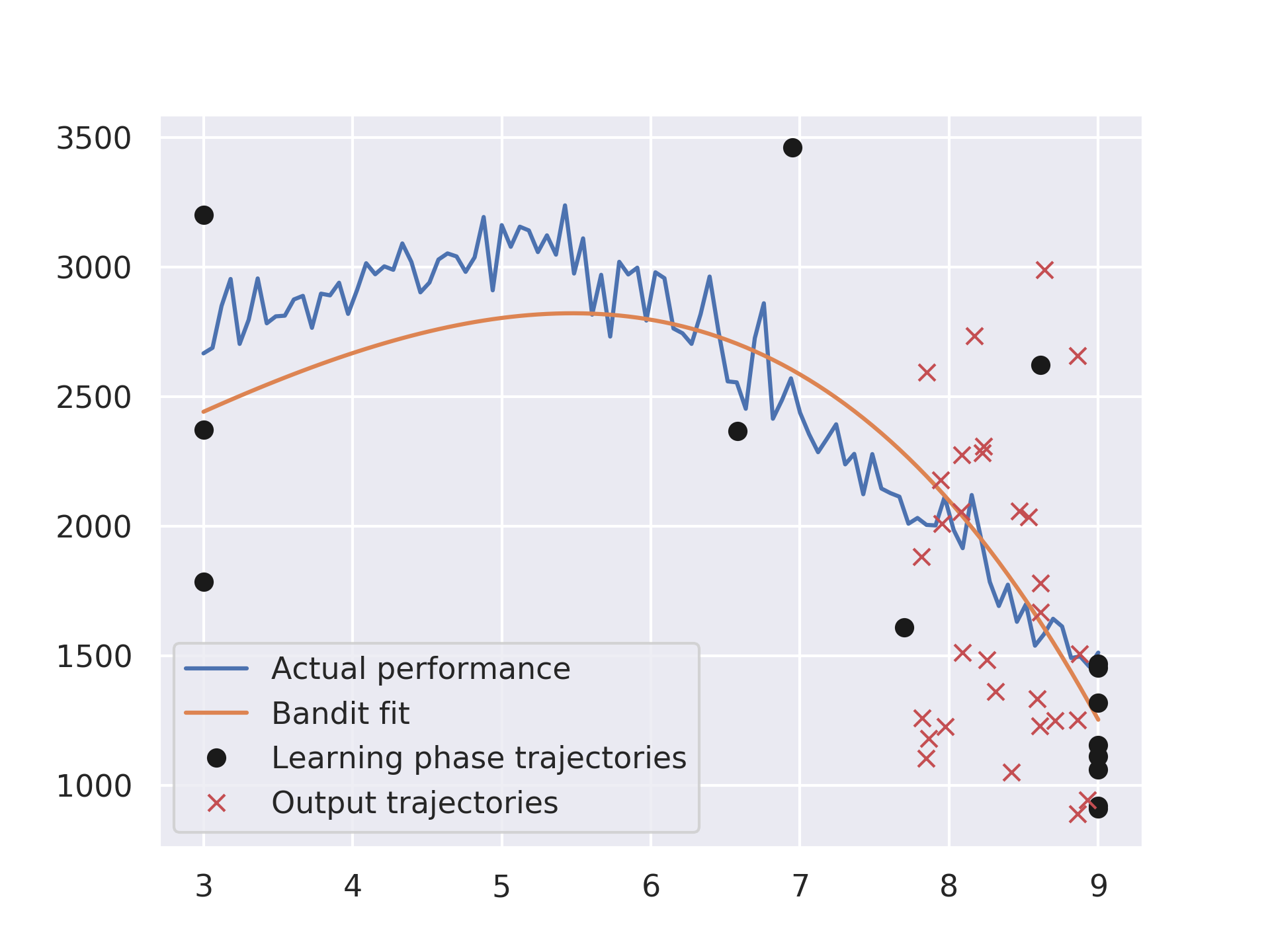}
\par\end{centering}
\caption{The bandit's operation in the 115th iteration of training in the Hopper
task. Individual trajectories are shown as dots or crosses, with the
position indicating the parameter value and the return obtained.}

\label{bandit-example}
\end{figure}

In Figure \ref{bandit-example}, we present the outcome from one particular
iteration of training. The true performance profile is estimated by
collecting 100 trajectories to calculate the mean return at each parameter
(again, these trajectories are not used for learning). Along with
this is shown the bandit's fit, the trajectories it collects while
learning and the trajectories that are sampled based on its estimate
of the bottom $\epsilon=0.1$ percentile (output used for training
the policy).

We see that the bandit takes exploratory actions (points to the left
and in the middle) that don't lead to the worst returns. However,
it quickly moves towards the region with low returns, and the final
fit is close to the true mean performance. From comparison with the
output trajectories, the trajectories in the learning phase are quite
clearly not representative of the bottom $\epsilon=0.1$ percentile
according to the source distribution. Thus, we cannot reuse these
to perform learning with the CVaR objective.

We also note that a perfectly learned performance profile is not necessary
to sample from the worst trajectories. As long as the fit is reasonably
accurate in that region, the output trajectories will be of good quality.
In practice, we expect LSB algorithms to be capable of doing this
as they tend to focus on these regions.

\subsection{(RQ2) Analysis of the Bandit Active Learner\label{subsec:Analysis-of-Bandit}}

Here we validate one of our key assumptions, that the active learner
learns well enough about the performance that it can produce a decent
approximation of a sample batch of the bottom $\epsilon$ percentile
of trajectories. For this, we first evaluate the average return for
a batch of samples from $\mathcal{P}$ by collecting a large number
of trajectories at each parameter. We note that these trajectories
are solely for the purpose of analysis and are not used to perform
any learning. Then, the percentile of the trajectory deemed to have
the greatest return among those chosen based on the bandit learner
is computed using the returns in this batch by using a nearest-neighbor
approximation. This is done across several iterations during the training,
and the median percentiles along with other statistics are reported
in Table \ref{pctl-table} for the Hopper task.

Ideally, we would like this value to come out to around $100\epsilon$
(when written as proper percentiles), i.e the parameter with the greatest
performance among the worst $\epsilon$ percentile should be at the
$\epsilon^{th}$ percentile. In our estimates, there are some outliers
that cause the average to become large, but as the median value shows,
it is indeed reasonably close to the desired value of 10 for $\epsilon=0.1$.

\subsection{(RQ3) Non-stationary Bandits for Data Reuse}

In this section, we attempt to answer RQ3 by considering modifications
that can be made to EffAcTS-C-B,
so that it uses lesser data, while also achieving a level of performance
and robustness that is close to the results above.

Particularly, we investigate the use of a modified version of the Thompson
Sampling algorithm above that is suited for non-stationary scenarios.
This is done in order to reuse performance history from previous iterations
to estimate the bandit's parameters with the aim of achieving a further
reduction in sample complexity. To implement this, past data is weighted
down in the Linear Regression step of Thompson Sampling, and the weight
decays after each iteration by a factor $\alpha$. That is, at the
$i^{th}$ iteration, the data from the $i^{th}$, $\left(i-1\right)^{th}$$\ldots$$\left(i-k\right)^{th}$
iterations would have weights 1, $\alpha$$\ldots$$\alpha^{k}$ respectively.

\begin{figure}[tb]
\begin{centering}
\includegraphics[width=1\columnwidth]{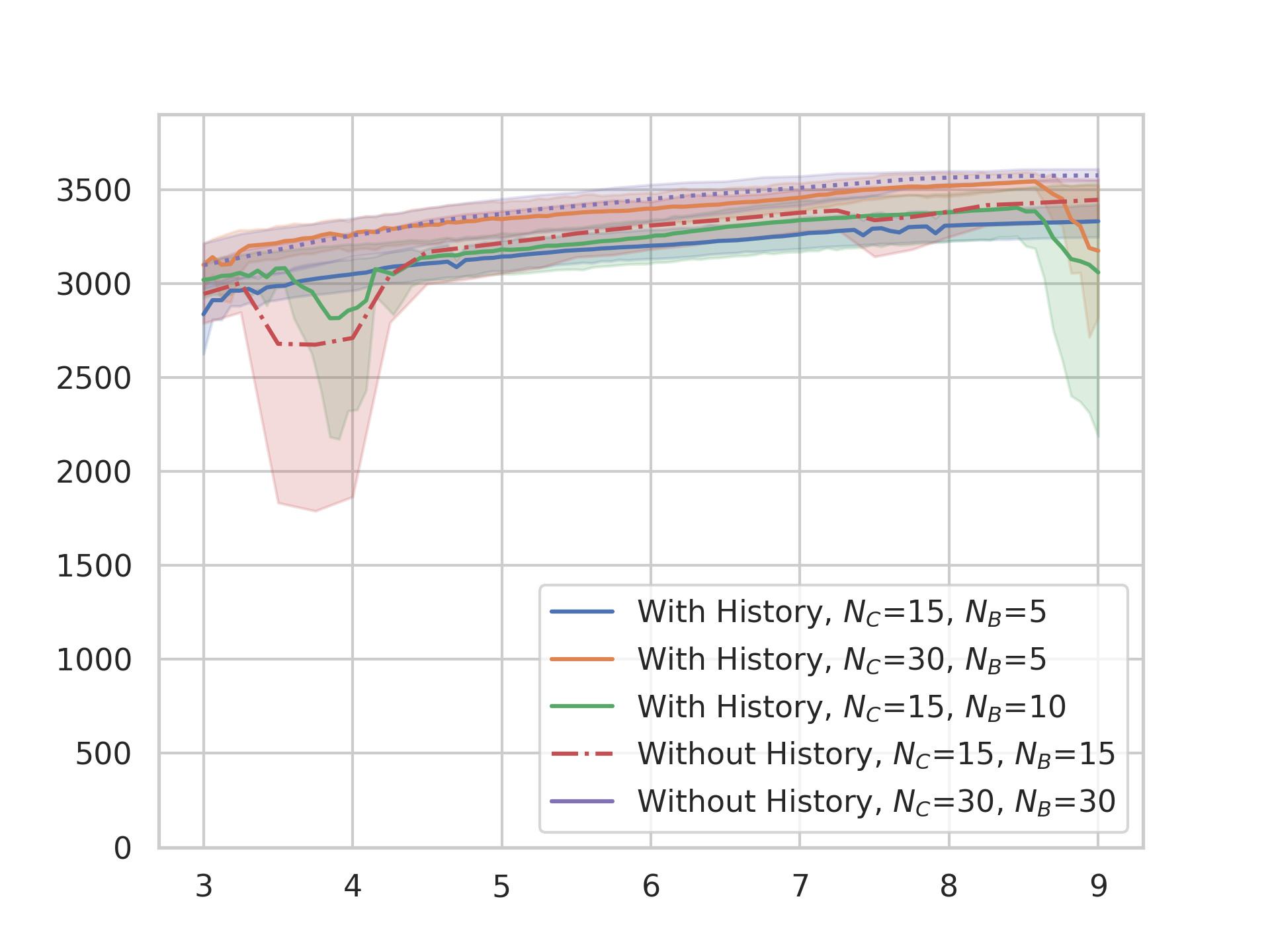}
\par\end{centering}
\caption{Results (performance vs. torso mass) for the non-stationary-bandit version (labeled ``With history'')
of EffAcTS-C-B, along with the original EffAcTS-C-B results (labeled
``Without history'') on the Hopper 1-D ensemble. A value of $\alpha=0.25$
was used, with the other hyperparameters remaining unchanged. As before,
the plot shows the performance as a function of torso mass for $\epsilon=0.1$,
and the bands indicate the confidence intervals of the performance
measured across 5 runs of the entire training procedure. This time, with 92\%
fewer trajectories than EPOpt, the configuration $N_{C}=15$, $N_{B}=5$ with history is able
to perform comparably to EPOpt, as well as EffAcTS without history and using more
trajectories.}

\label{hist-res}
\end{figure}

We compare this version with the original setup in Figure \ref{hist-res}.
We see that the setting with the smallest number of trajectories collected
with history ($N_{C}=15$, $N_{B}=5$) is more robust than the vanilla
case with $N_{C}=15$, $N_{B}=15$ and performs nearly as well (despite
cutting down on 10 more trajectories). With more data, some loss of
robustness is encountered, but the performance improves, and becomes
comparable to the best original case ($N_{C}=30$, $N_{B}=30$).

\subsection{Other Remarks}

We note that we do not perform ``pre-training'' as in EPOpt where
the entire batch of trajectories is used for policy learning for some
iterations at the beginning (corresponding to optimizing for the average
return over the ensemble). This has been reported to be necessary,
possibly due to problems with initial exploration if using the CVaR
objective from the beginning. EffAcTS-C-B on the other hand works
without any such step. However, at the very beginning, we collect
trajectories for parameters sampled directly from $\mathcal{P}$ until
2048 time steps have elapsed in that iteration. This is because algorithms
like TRPO have been known to require at least that much data per iteration.

\section{Conclusions and Further Possibilities}

We developed the EffAcTS framework for using active learning to make
an informed selection of model parameters based on agent performance,
which can subsequently be used to judiciously generate trajectories
for robust RL. With an illustration of this framework based on Linear
Bandits and the CVaR objective, we have both demonstrated its applicability
for robust policy search as well as established its effectiveness
in reducing sample complexity by way of empirical evaluations on standard
continuous control domains. We also discussed our work in the context
of Multi-Task Learning along with the similarities and differences
between these settings.

Our work opens up requirements for active learning algorithms that
can work well with even lesser data than we need here. Methods like
Gaussian Process Regression are known to be efficient, but not in
high dimensional spaces. For robust policy search methods to be effective
for transfer from simulation to reality, they need to be able to handle
the complexities of the real world, which necessitates methods that
work with high dimensional model ensembles, which in turn entail frameworks
such as EffAcTS to help reduce the sample complexity. Another possibility
for robust policy search itself is to develop objectives that can
speedup learning as well as make use of the features of EffAcTS to
maintain sample efficiency.

With our interpretation of robust policy search as a parameterized
version of Multi-Task Learning, a natural next step would be to adapt
developments in the usual discrete MTL setting to robust policy search.
It would also be worthwhile to similarly investigate the applicability
of Meta Learning, as it would prove useful for both dealing with large
disparities between the source domains and the real world, as well
as coping with unmodeled dynamics (which are unavoidable since it
is not feasible to model the real world with complete accuracy).

\section{Acknowledgments}

The authors thank the Robert Bosch Center for Data Science and AI,
IIT Madras for funding this work
and providing the requisite computing resources. We also thank Aravind
Rajeswaran for valuable pointers and suggestions, and OpenAI for their
excellent codebase of Deep RL algorithms.

\bibliographystyle{ACM-Reference-Format}
\bibliography{refs}

\end{document}